\documentclass[a4paper,fleqn]{cas-sc}

\usepackage[authoryear,longnamesfirst]{natbib}

\def\tsc#1{\csdef{#1}{\textsc{\lowercase{#1}}\xspace}}
\tsc{WGM}
\tsc{QE}
\tsc{EP}
\tsc{PMS}
\tsc{BEC}
\tsc{DE}
\begin{document}
\let\WriteBookmarks\relax
\def\floatpagepagefraction{1}
\def\textpagefraction{.001}
\shorttitle{Image Manipultation with Natural Language}
\shortauthors{D.Zhu, A.Mogadala and D.Klakow}
%\begin{frontmatter}

\title [mode = title ]{Image Manipulation with Natural Language using Two-sided Attentive Conditional Generative Adversarial Network} 
\author{Dawei Zhu}
\cormark[1]
\author{Aditya Mogadala}
\cormark[1]
\author{Dietrich Klakow}
\address{Spoken Language Systems (LSV), Saarland University, Sarlaand Informatics Campus, Saarbr{\"u}cken, Germany}
\cortext[cor1]{Contributed equally}
\begin{abstract}
Altering the content of an image with photo editing tools is a tedious task for an inexperienced user. Especially, when modifying the visual attributes of a specific object in an image without affecting other constituents such as background etc. To simplify the process of image manipulation and to provide more control to users, it is better to utilize a simpler interface like natural language. Therefore, in this paper, we address the challenge of manipulating images using natural language description. We propose the Two-sidEd Attentive conditional Generative Adversarial Network (TEA-cGAN) to generate semantically manipulated images while preserving other contents such as background intact. TEA-cGAN uses fine-grained attention both in the generator and discriminator of Generative Adversarial Network (GAN) based framework at different scales. Experimental results show that TEA-cGAN which generates 128 $\times$ 128 and 256 $\times$ 256 resolution images outperforms existing methods on CUB and Oxford-102 datasets both quantitatively and qualitatively.
\end{abstract}

\maketitle

\section{Introduction}
\label{sec:intro}
Large availability of image capturing devices and storage platforms has increased end-users interest in creation and storage of images. These platforms also help users to share their images allowing modifications that match their needs such as making them look better. Consequently, the demand for manipulating images online or offline is increasing. However, manipulation of images is non-trivial for non-experts who do not have apprehension about underlying principles of both photo-editing tools and image processing techniques. 

To simplify the process of image manipulation, automatically changing various aspects of images is an interesting direction to explore. In earlier research, different ways of automatic image manipulation are examined . Initially, approaches transformed grayscale into a color image~\cite{zhang:2016} or transferred their style~\cite{gatys:2016} matching numerous well-known artworks. Few approaches~\cite{liang:2017} took the desired object category in an image as input and then learned to change the object by modifying their appearance or geometric structure. There has been also another direction of interest~\cite{zhu:2016} shown to manipulate images by projecting them onto an image manifold with various user's scribbles as input. It has been further extended to handle various domains in the context of paired~\cite{isola:2017} and unpaired~\cite{zhuunpair:2017} image-to-image translation without hand-engineering loss functions. There also exist some more variations of image manipulation, more details can be referred from the recent surveys~\cite{mogadala:2019}. 

\begin{figure*}
    \centering
        \includegraphics[width=0.9\textwidth]{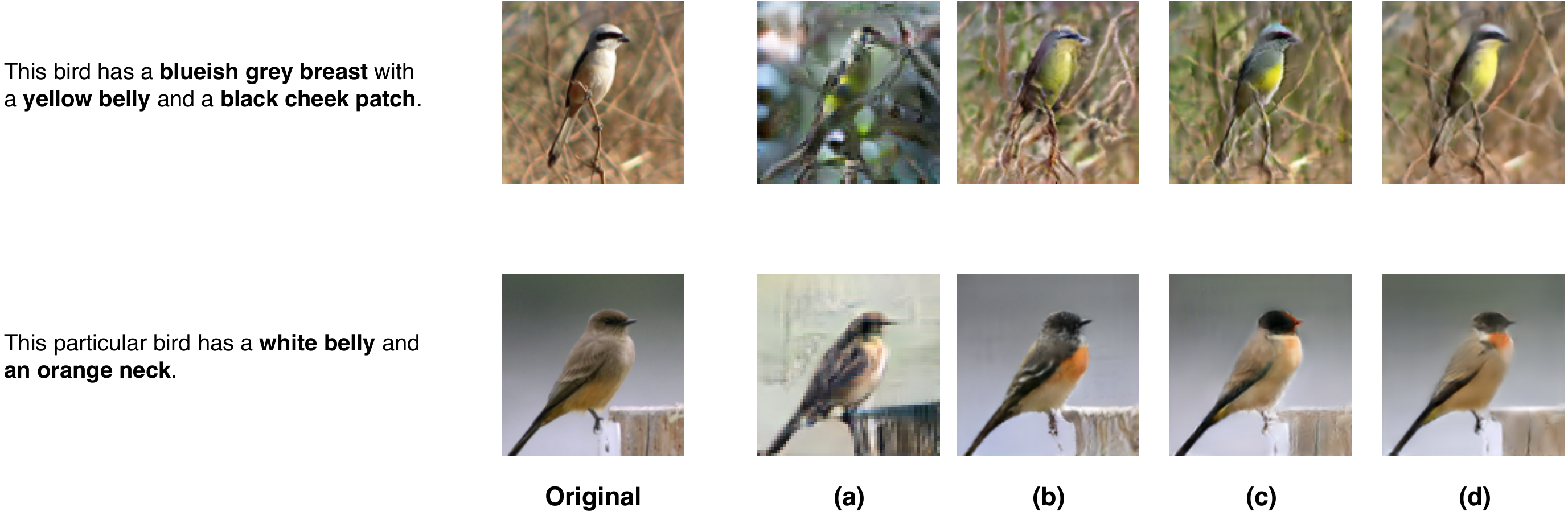}
    \caption{Samples of the output images by manipulating input images using natural language description. Existing methods (a) SISGAN~\cite{dong:2017} and (b) TAGAN~\cite{nam:2018} produce reasonable results, but fail to preserve text-irrelevant contents. Our methods (c) Single-scale and (d) Multi-scale accurately manipulates images according to the text while preserving the background.}\label{fig:immanip}
\end{figure*}

Although earlier mentioned research has achieved promising results, manipulation of images according to user intention becomes more difficult as those methods allow minimal or no control on the image generation. To address it, recent approaches have leveraged text-to-image generation~\cite{reed:2016} techniques and manipulated images using natural language description~\cite{dong:2017,nam:2018}. To be specific, the focus is on modifying visual attributes of an object in an image, where the visual attributes are characterized by the color and the texture of an object. Figure~\ref{fig:immanip} outline the goal of the task by showing samples of the manipulated images using different models.
Nevertheless, current natural language based image manipulation approaches still suffer from several problems. 
\begin{itemize}
    \item First, they fail to properly attend on locations that need to be modified and alter basic structure in the image (e.g., layout and pose of the object) during generation.
    \item Second, they generate lower resolution images (e.g., 64 $\times$ 64 or 128 $\times$ 128), while recent works~\cite{karras:2017} show that higher resolution images (e.g., 256 $\times$ 256) are preferred by humans due to their improved quality and stability.
\end{itemize}

Therefore, in this paper, we overcome the limitations of earlier approaches by proposing a novel Two-sided Attentive Conditional Generative Adversarial Network (TEA-cGAN) for generating manipulated images while preserving other contents such as the background. Particularly in our generator, we compute a matching score between every sub-region in an image and the word in the natural language description. To be specific, a \textit{attention map} consisting of matching scores is constructed to be used along with the image input features. This helps local word-level features (i.e., fine-grained) to attend a specific type of visual attribute and detach text-relevant areas in the image from irrelevant ones. Similarly, the discriminator decides whether the image is real or fake by accumulating fine-grained matching scores with attention. Using the feedback from the discriminator, our generator adapts itself to generate manipulated images.

To the best of our knowledge, none of the previous works propose attention over conditional Generative Adversarial Network (cGAN) in a generator for fine-grained image manipulation with natural language. However, attention has been used for text-to-image generation~\cite{xu:2018} or applied only to the discriminator~\cite{nam:2018}. 

The summary of the main contributions in this work is listed as follows.

\begin{itemize}
    \item We proposed the novel architecture TEA-cGAN for image manipulation with natural language by leveraging fine-grained attention on conditional GAN both in the generator and discriminator.
    \item We have built the generator with two different scales to support the generation of sharper and higher resolution images.
    \item We thoroughly evaluate our approach on two datasets containing a different type of images.
\end{itemize}

\section{Two-sided Attentive Conditional Generative Adversarial Network (TEA-cGAN)}
\label{sec:attcgan}
Let $\mathbf{I}$, $\mathbf{T}$, $\mathbf{\hat{T}}$ denote an image, a positive natural language description matching the image, and a mismatch description that does not correctly describe the image, respectively. Given an image $\mathbf{I}$ and a target mismatch text $\mathbf{\hat{T}}$, our aim is to manipulate $\mathbf{I}$ according to $\mathbf{\hat{T}}$ so that the visual attributes of the manipulated image $\mathbf{\hat{I}}$ match the description of $\mathbf{\hat{T}}$ while preserving other information (e.g., background). We use Generative Adversarial Network (GAN)~\cite{goodfellow:2014} as our framework, in which the generator is trained to produce $\mathbf{\hat{I}}$ given $G(\mathbf{I},\mathbf{\hat{T}})$. In the following, we describe the generator, discriminator, and objective of our TEA-cGAN in detail.

\subsection{Generator}
\label{ssec:generator}
The generator is an encoder-decoder architecture with attention inspired by the plain text-to-image generation approach~\cite{xu:2018}. We design two different variants of it (i) Single-scale and (ii) Multi-scale. In the following, details about each of them are presented separately.

\subsubsection{Single-scale }
\label{ssec:ssfa}
\begin{figure*}
    \centering
        \includegraphics[width=\textwidth]{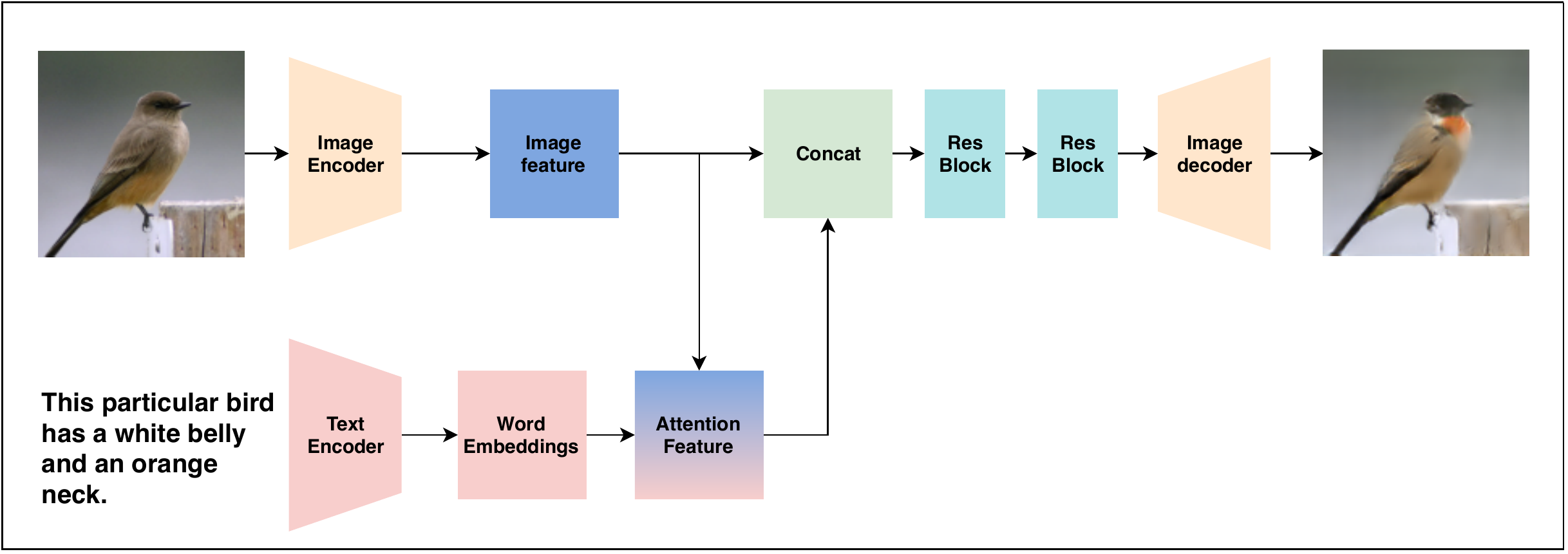}
    \caption{The architecture of our Single-scale generator. It has an encoder-decoder structure, the text is injected into the image generation process using an attention mechanism.}\label{fig:singlescale}
\end{figure*}

We first encode the input image to a feature representation with an image encoder. Further, only the final output representation of the image encoder is used in combination with the fine-grain word-level features arising from the natural language description. This is done to focus and modify only the text-relevant regions in an image and leaving other regions untouched. The structure of our Single-scale model is shown in the Figure~\ref{fig:singlescale}.

Image encoder is designed using a convolutional network which has a representation of $\mathbf{I}$ defined by a tensor $V = (v_1 \ldots v_N) \in \mathbb{R}^{M \times N}$, where $\text{M}$ denotes the number of feature maps and $\text{N}$ denotes the spatial dimension \footnote{A spatial dimension normally consists of a height $H$ and a width $W$, we use $N=H\times W$ for simple notation}. The natural language description encoder generates a representation of $\mathbf{\hat{T}}$ (or $\mathbf{T}$), denoted by $W \in \mathbb{R}^{D \times L}$, using a bidirectional long short-term memory (LSTM)~\cite{hochreiter:1997} where $w_i \in \mathbb{R}^D$ represents the $i$-th word in the description obtained by concatenating forward and backward LSTM hidden layers. $L$ is the length of the description.

We jointly process both $V$ and $W$ to compute a matching score between $v_i$ and $w_j$ for generating different sub-regions of the image conditioned on the relevant words corresponding to those sub-regions. 
\begin{equation}
w'_{j} = \mathcal{U}w_j
\end{equation}
Where  $w'_{j}$ represents a bilinear projection of $w_j$ into the $v_i$ space with a projection matrix $\mathcal{U} \in \mathbb{R}^{N \times D}$. Further, we use $w'_{j}$ to make the final classification decision by adding word-level attentions to reduce the impact of less relevant words. Our attention is a Softmax
across $L$ words and is computed by Equation~\ref{eqn:attn}. Word-context feature ($v'_{i}$) is computed by its linear combination with $w'_{j}$ given by Equation~\ref{eqn:weighattn}.
\begin{equation}
\label{eqn:attn}
    \alpha_{ij} = \frac{exp(v_{i}^Tw'_{j})}{\sum_{k=1}^{L} exp(v_{ik}^Tw'_{jk}) }
\end{equation}
\begin{equation}
\label{eqn:weighattn}
v'_{i} = \sum_{j=1}^{L} \alpha_{ij} w'_{j}
\end{equation}

Here, $V'  = (v'_{1},\ldots,v'_{N})$ denote the word-context features of the entire image. Finally, $\mathcal{V}$ and $\mathcal{V'}$ will be concatenated and fed into several residual blocks for further processing. The processed feature will be further transformed into an image using the image decoder.

\subsubsection{Multi-scale }
\label{ssec:msfa}

\begin{figure*}
    \centering
        \includegraphics[width=\textwidth]{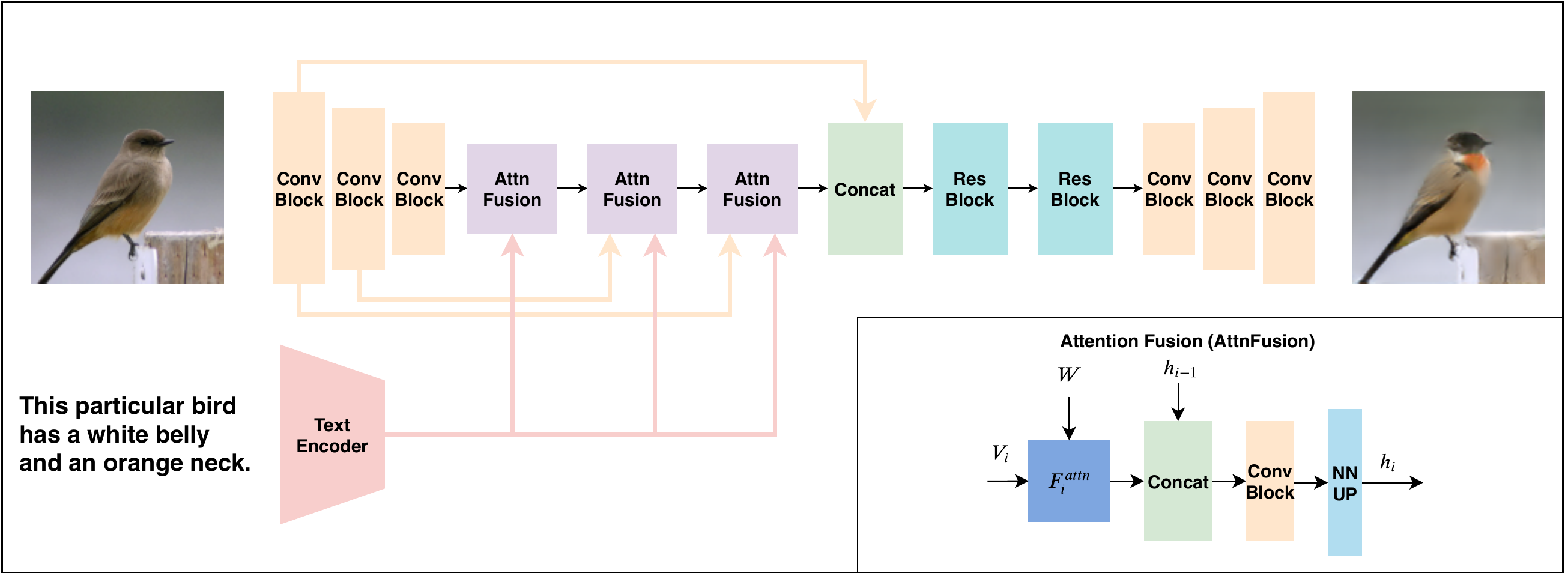}
    \caption{The  architecture  of  our  Multi-scale generator. We utilize image features of different scales while generating images. For each scale, the text information is injected into the network by an attention mechanism as shown in \textit{AttnFusion}.}\label{fig:multiscale}
\end{figure*}

The main issue with the Single-scale architecture is that it may fail to focus on the right locations in images. We understand that $v_i$ encodes the information of a sub-region in an image and the size of it depends on the receptive field of the last convolutional layer. If the receptive field is too small, $v_i$ may fail to provide necessary information for computing a matching score used for the attention. However, if the receptive field is too large, the corresponding sub-regions may contain irrelevant features for the attention. Considering an extreme case that the size of the receptive field equals the size of the image, then the attention does not make much sense. Hence, we are motivated to utilize features from different scales. The structure of our Multi-scale model is shown in the Figure~\ref{fig:multiscale}.

To be specific, we use a set of tensors $(V_m,\ldots, V_1)$ \footnote{we intentionally use the reversed order of indexing for simple notation in the following equations.} constituting image $\mathbf{I}$ features of different scales extracted by the different convolutional layers and a embedding matrix $W$ consisting of word embeddings in the natural language description (the same $W$ as defined in the single-scale model). We starts by computing the word-context feature $\mathcal{F}^{attn}(V_1)$ using the image feature ($V_1$) obtained by the last convolutional layer. This is done in the same manner as the Single-scale model. It is further upsampled spatially using nearest neighbor upsampling. The resulting feature is treated as our first hidden state $h_0$. Furthermore, we develop an \textit{attention fusion} (AttnFusion) module to fuse the attention information from different scales for generating new hidden states. The structure of the \textit{attention fusion} module is shown in the sub-box of Figure~\ref{fig:multiscale}. Specifically, the hidden states ($h_i$) are computed as follows:

\begin{equation}
h_{0} = NN_{\uparrow}(\mathcal{F}^{attn}(V_1,\mathbf{W}))
\end{equation}
\begin{equation}
\begin{split}
h_{i} &=AttnFusion(h_{i-1}, V_{i}, W) \\
      &=NN_{\uparrow}(Conv(\mathcal{F}^{attn}(V_i, W) \circ h_{i-1}))
\end{split}
\end{equation}

Where $NN_{\uparrow}$ denotes the nearest neighbour upsampling and $\circ$ represents the spatial concatenation between two feature tensors. Note $\mathcal{F}^{attn}$ needs to be adapted accordingly to fit the spatial dimension of $V_i$. Further, the final hidden state is concatenated with visual features and they are fed into residual blocks for further processing. Finally, an image decoder will generate the processed feature back to an image. 

Architecture-wise the convolutional operations in the generator do not have biases and we do not apply batchnorm to the output layer of the generator. Also, in the Multi-scale generator, we move the last convolutional layer in the residual blocks to the second layer in the image encoder as it improves the quality of image generation.

\subsection{Discriminator}
\label{ssec:discriminator}
The main aim of our research is to incorporate attention into the generator. Hence, for both the Single-scale and the Multi-scale model we apply a discriminator in line with Nam et al.~\cite{nam:2018}. However, we do fine-tuning of the discriminator. Similarly to the generator, all convolutional operations do not have biases and we do not apply batchnorm to the discriminator input layer. We use the conditional and unconditional score in discriminator, where the conditional score is influenced by both the input image and the description while the unconditional score is only influenced by the input image.

\subsection{Objective}
\label{ssec:objective}
Let $D(I,T)$ denote the output (score) of the discriminator by considering both the image quality of image $I$ and the matching between $I$ and a text description $T$. Higher score indicates higher image quality and a good matching \footnote{Generally, the score is normalized to $[0,1]$}. In addition, inspired by \cite{zhang:2018}, we also introduce the unconditional score $D(I)$ to purely assess the image quality. We use a factor $\gamma_1$ to balance the influence between the conditional and unconditional loss. The discriminator's objective $\mathcal{L}_D$ is defined as follows:
\begin{equation}
\label{eqn:objdisc}
\begin{split}
\mathcal{L}_D &= \mathbb{E}_{\mathbf{I} \sim p_{data}} [\log{D(\mathbf{I})}] \\
 &+ \mathbb{E}_{\mathbf{I} \sim p_{data}}[\log{(1 - D(G(\mathbf{I})))}] \\
 &+\mathbb{E}_{\mathbf{I}, \mathbf{T} \sim p_{data}} [\gamma_1 \log{D(\mathbf{I},\mathbf{T})}] \\
 &+ \mathbb{E}_{\mathbf{I}, \mathbf{\hat{T}} \sim p_{data}} [\gamma_1 \log{(1 - D(G(\mathbf{I},\mathbf{\hat{T}}),\mathbf{\hat{T}}))}]
\end{split}
\end{equation}
The mismatch text $\mathbf{\hat{T}}$ is randomly sampled from a dataset regardless of $\mathbf{I}$. Similar to text-to-image task, we feed both positive and negative examples (($\mathbf{I},\mathbf{T})$ and $G(\mathbf{I},\mathbf{\hat{T}})$, respectively) to the discriminator to make it not only judge image quality but also the matching. Note that we want to maximize this objective. In training, we minimize $-\mathcal{L}_D$.
Further, the generator objective ($\mathcal{L}_G$) is defined as follows:
\begin{equation}
\begin{split}
\mathcal{L}_G & = \mathbb{E}_{\mathbf{I} \sim p_{data}}[\log{D(\mathbf{I})}] \\
&+ \mathbb{E}_{\mathbf{I}, \mathbf{\hat{T}} \sim p_{data}}[\gamma_1 \log{D(G(\mathbf{\hat{I}},\mathbf{\hat{T}}),\mathbf{\hat{T}})}]\\
&+ \gamma_2 L_{R}
\end{split}
\end{equation}
where $\gamma_2$ is a hyperparameter that controls the influence of the auxiliary losses. We wish the generator to keep the background intact while manipulating the image as descriptions only target to the main objects in the images. In other words, we want the generator to do minimum change to the input images. Hence, we add the image reconstruction loss $L_R$. For positive pairs, i.e. an image $I$ and a matched text $T$ the generator should reconstruct $I$, changes on the original image will be penalized. We use the $L_1$ loss  (Equation~\ref{eqn:l1}) in training.
\begin{equation}
\label{eqn:l1}
L_{R} = \lVert \textbf{I} - G(\textbf{I},\textbf{T}) \rVert
\end{equation}
Our TEA-cGAN is trained by alternatively minimizing both the discriminator and the generator objectives.

\section{Experimental Setup}
\subsection{Datasets}
We evaluated our approaches by conducting experiments with the modified versions of the Caltech-200 bird dataset (CUB)~\cite{wah:2011} and Oxford-102 flower dataset~\cite{nilsback:2008}. Each image in the original dataset was crowdsourced by~\cite{reedlearning:2016} to collect 10 captions for describing the colors of different parts of birds or flower. More details about the datasets are given in the Table~\ref{table:oxford-102-dataset} and Table~\ref{table:cub-dataset}.
\begin{table}
\small
    \centering
     \caption{\label{table:oxford-102-dataset} Splits of ``Oxford-102'' dataset with image descriptions.}
    \begin{tabular}{l | c c c c}
    \toprule
         Split          & Images    & Captions per Image    & Classes \\
    \midrule
        Training        &5,878      & 10                    & 82 \\
        Validation      &1,156      & 10                    & 20 \\
        Test            & 1,155     & 10                    & 20 \\
        \midrule
        Total           & 8,189     &10                     & 102 \\
    \bottomrule
    \end{tabular}
\end{table}

\begin{table}
\small
    \centering
    \caption{ \label{table:cub-dataset} Splits of ``CUB'' dataset with image descriptions.}
    \begin{tabular}{l | c c c}
    \toprule
        Split           & Images    & Captions per Image    & Classes \\
    \midrule
        Training        & 8,855     & 10                    & 150 \\
       % Validation      & - & - & - \\
        Test            & 2,933     & 10                    & 50\\
    \midrule
        Total           & 11,788    & 10                    & 200\\
    \bottomrule
    \end{tabular}
\end{table}

\subsection{Evaluation}
Automatic evaluation of GAN approaches is tricky. Although Inception Score~\cite{salimans:2016} and Fr{\'e}chet Inception distance (FID)~\cite{heusel:2017} are used as a quantitative measure for evaluation of generated images, it cannot be used for our case since generated images do not have any ground truth labels (more detail in section \ref{subsec: inception}). Similar to earlier approaches~\cite{dong:2017,nam:2018},  we will conduct a human evaluation to rank our proposed approaches and existing models on two different aspects:
\begin{itemize}
    \item \textit{Accuracy}: Do the generated images match the description while preserving the background of the input image?
    \item \textit{Naturalness}: Do the generated images look realistic?
\end{itemize}

We also calculate the reconstruction losses (L$_1$ and L$_2$) per pixel to determine the error attained in reconstructing back the input image while keeping the background intact. 

\subsection{Implementation}
\label{ssec:impl}
We implemented our proposed approaches using PyTorch 1.1.0\footnote{\url{https://pytorch.org/}}. Initially, word embeddings of the natural language descriptions are initialized with the fastText\footnote{\url{https://github.com/facebookresearch/fastText}} vectors and data augmentation is applied to the input images by random flipping and cropping. 
The weight $\gamma_2$ of the reconstruction loss is set to 2 for Single-scale generator and 3 for the Multi-scale generator (for all resolutions), while weight $\gamma_1$ is set to 10. A batch-size of 128 is used for generating images of resolution of $128 \times 128$, and a batch-size of 32 is used for generating images with a resolution of $256 \times 256$. We trained all our models for 600 epochs with an Adam Optimizer~\cite{kingma:2014} having a learning rate of 0.0002 and a momentum of 0.5. The learning rate is decayed by 0.5 for every 100 epochs.

\section{Experimental Results}
\label{sec:exp}
We conducted experiments at different levels to evaluate our proposed models. First, a quantitative analysis is performed by calculating reconstruction losses. To further validate our results, we conducted a human study on the ``CUB'' dataset to comprehend  \textit{Accuracy} and \textit{Naturalness} of generated images by ranking the best models. We then provide qualitative results of the generated images comparing different methods\footnote{We use 128 $\times$ 128 to have a fair comparison with other methods.} and also higher resolution images from our Multi-scale model. Later, we show visualizations of the attention for Multi-scale model and further analyze the impact of text interpolation and the contributions from components of the model.

\subsection{Quantitative Results}
\label{ssec:quantres}
To compare the model's ability to keep the text irrelevant content preserved (e.g., background), we first calculate the reconstruction loss using the image along with their natural language description from both ``CUB'' and ``Oxford-102'' datasets. In the Table~\ref{table:recoeval}, we see our TEA-cGAN-Multi-Scale model has the lowest reconstruction losses (L$_1$ and L$_2$) indicating that our Multi-scale model is the preferred choice for keeping the content of the original image intact.
\begin{table*}
  \centering
  \caption{\label{table:recoeval} L$_1$ and L$_2$ Loss (pixel-level) on CUB and Oxford-102 test dataset. Lower the better.}
  \begin{tabular}{lcccc}
    \toprule
    & \multicolumn{2}{c}{CUB} & \multicolumn{2}{c}{Oxford-102} \\
    \cmidrule(r){2-3}
    \cmidrule(r){4-5}
    Method & L$_1$ & L$_2$ & L$_1$ & L$_2$ \\
    \cmidrule(r){1-1}
    \cmidrule(r){2-3}
    \cmidrule(r){4-5}
    SISGAN~\cite{dong:2017} & 0.51 & 0.20  & 0.53 &  0.19 \\
    TAGAN~\cite{nam:2018} & 0.46 & 0.15  & 0.48 &  0.16\\
    \midrule
    TEA-cGAN-Single-Scale (Ours) & 0.37  & 0.1  & 0.5 & 0.17 \\
    TEA-cGAN-Multi-Scale (Ours) & \textbf{0.25}  & \textbf{0.05} & \textbf{0.33} & \textbf{0.07}  \\
    \bottomrule
  \end{tabular}
\end{table*}

We then perform a human evaluation on the CUB dataset by ranking images generated by different models based on (i) Accuracy and (ii) Naturalness. For the evaluation, we randomly selected 8 images and 8 texts from the test set and produced 64
outputs from the CUB dataset for each method. We resized all output images to 128 $\times$128 to have a fair comparison and prevent the users from evaluating the images based on different resolutions. Table~\ref{table:quanteval} shows the results from the study where results are shown as average ranking values.
\begin{table*}
  \centering
  \caption{\label{table:quanteval} Accuracy and Naturalness average ranking values evaluated by users on CUB test dataset. Lower the better.}
  \begin{tabular}{lcc}
    \toprule
    & \multicolumn{2}{c}{CUB} \\
    \cmidrule(r){2-3}
    Method & Accuracy & Naturalness \\
    \cmidrule(r){1-1}
    \cmidrule(r){2-3}
    SISGAN~\cite{dong:2017} & 3.94  & 3.97 \\
    TAGAN~\cite{nam:2018} & 2.4  & 2.6 \\
    \midrule
    TEA-cGAN-Single-Scale (Ours) & 1.97  & 1.94  \\
    TEA-cGAN-Multi-Scale (Ours) & \textbf{1.69}  & \textbf{1.49} \\
    \bottomrule
  \end{tabular}
\end{table*}

To further verify the study results, we conducted a Chi-Square test. We want to test whether the models and the ranking are dependent. The null hypothesis is that these two variables are independent. We compared our TEA-cGAN-Multi-scale model with all other models. We also compare our TEA-cGAN-Single-scale model with the SISGAN~\cite{dong:2017} and TAGAN~\cite{nam:2018}. Accuracy and naturalness are tested separately. In total, we conducted 10 significance tests and all of them get a $p$-value smaller than $10^{-18}$. Even if we use Bonferroni correction the $p$-value is much smaller than $5\%$ thus we reject the null hypothesis. Table \ref{table:Chi_acc} and Table~\ref{table:Chi_natural} presents all $p$-values computed with all models used for evaluation i.e., SISGAN, TAGAN, TEA-cGAN-Single-scale (Single-scale) and TEA-cGAN-Multi-scale (Multi-scale).

\begin{table}
\caption{\label{table:Chi_acc} Cross comparison the models. Accuracy.} % title of Table
\centering % used for centering table
\begin{tabular}{lc} % centered columns (4 columns)
\toprule %inserts double horizontal lines
Model & Accuracy  \\ % inserts table
      & ($p$-value) \\
%heading
\midrule  % inserts single horizontal line
SISGAN \& Multi-scale & $2.49 \times 10^{-148}$  \\ % inserting body of the table
TAGAN \& Multi-scale & $3.95 \times 10^{-29}$  \\
Single-scale \& Multi-scale & $2.77 \times 10^{-11}$  \\
SISGAN \& Single-scale & $2.04 \times 10^{-151}$   \\
TAGAN \& Single-scale & $1.14 \times 10^{-19}$  \\% [1ex] adds vertical space
\bottomrule %inserts single line
\end{tabular}
\end{table}

\begin{table}
\caption{\label{table:Chi_natural} Cross comparison the models. Naturalness.} % title of Table
\centering % used for centering table
\begin{tabular}{lc} % centered columns (4 columns)
\toprule %inserts double horizontal lines
Model & Naturalness \\  % inserts table
      & ($p$-value)\\
%heading
\midrule  % inserts single horizontal line
SISGAN \& Multi-scale& $1.22 \times 10^{-159}$  \\ % inserting body of the table
TAGAN \& Multi-scale  & $1.52 \times 10^{-69}$ \\
Single-scale \& Multi-scale & $6.85 \times 10^{-16} $  \\
SISGAN \& Single-scale  & $2.44\times 10^{-156}$  \\
TAGAN \& Single-scale  & $1.17\times 10^{-32}$ \\ % [1ex] adds vertical space
\bottomrule %inserts single line
\end{tabular}
\end{table}

\subsection{Qualitative Results}
\label{ssec:qualres}
\begin{figure}
    \centering
        \includegraphics[width=0.75\textwidth]{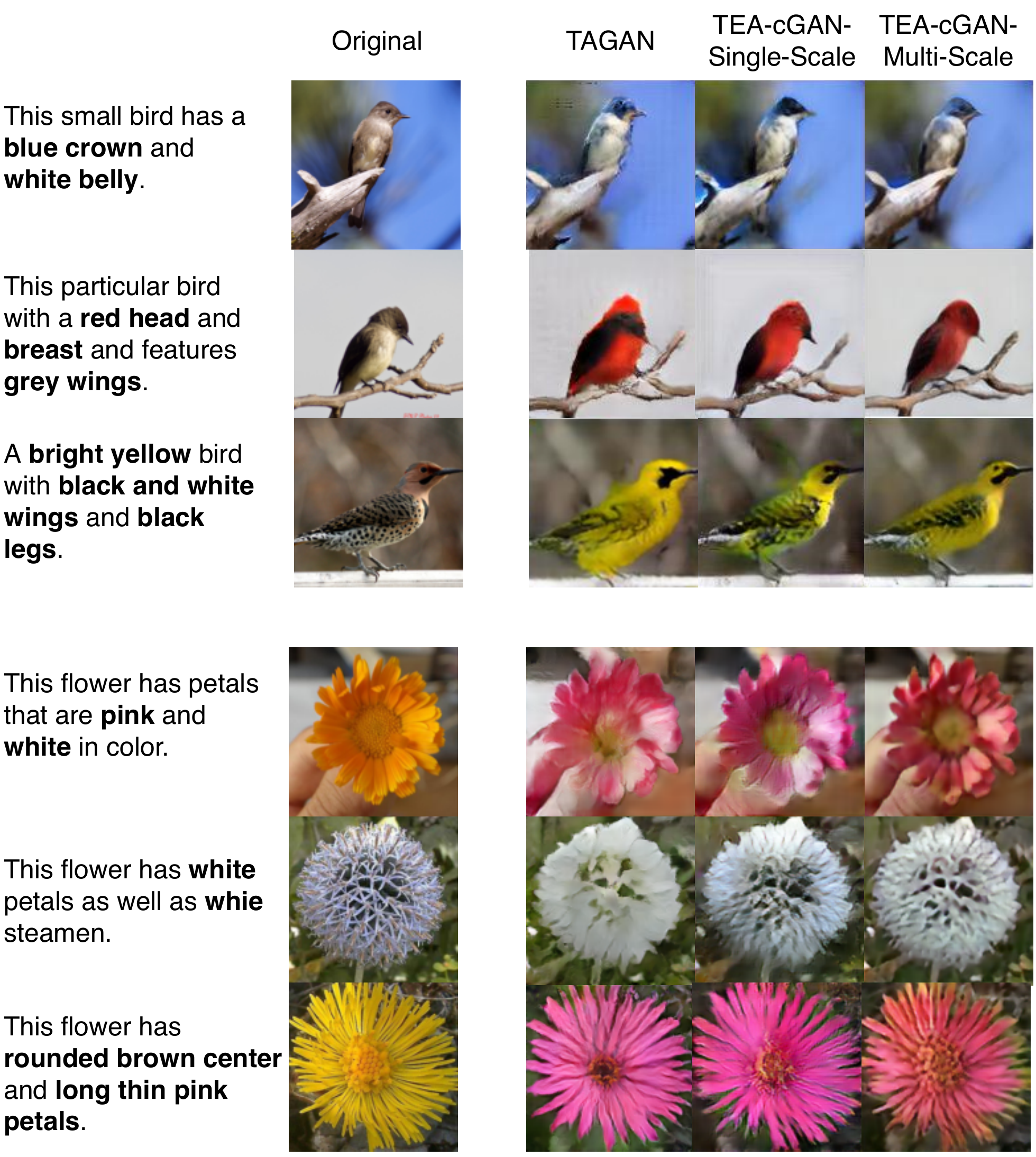}
    \caption{Qualitative comparison of our proposed methods with the closest baseline  TAGAN~\cite{nam:2018}. In many cases, our proposed TEA-cGAN-Multi-scale method outperforms the baseline methods qualitatively.}\label{fig:qualcomp}
\end{figure}

Figure~\ref{fig:qualcomp} shows a qualitative comparison of our models with a strong baseline (i.e., TAGAN) that generates a 128 $\times$ 128 resolution. We observe that the baseline can generate an image matching the natural language descriptions. However, the method fails to keep the content relevant to the text and is likely to generate a different image in contrast to the original layout. However, our method preserves the background intact and helps in only transferring visual attributes given in the text.

\begin{figure}
    \centering
        \includegraphics[width=0.75\textwidth]{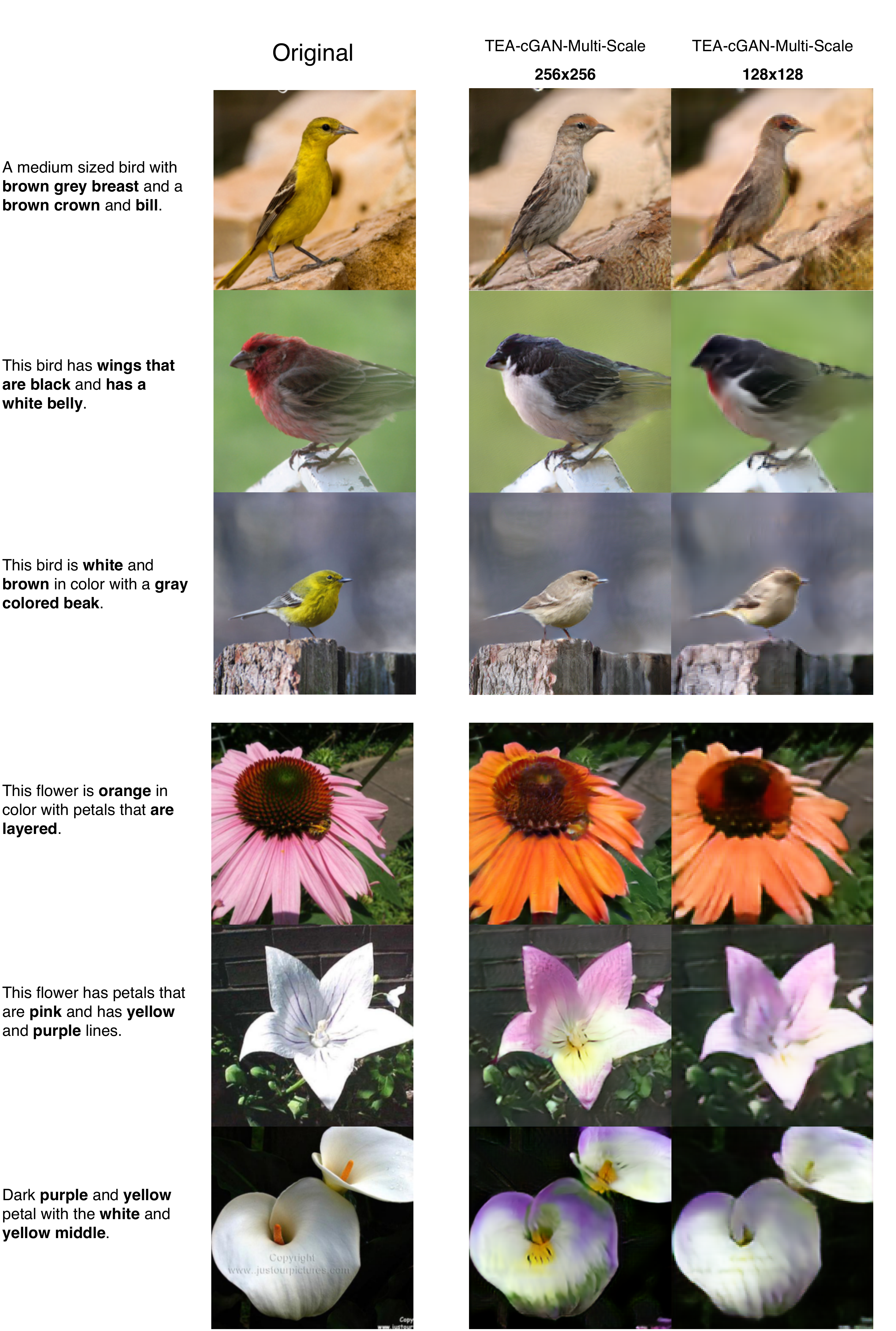}
    \caption{Samples of the  128 $\times$ 128 and 256 $\times$ 256 resolution images generated with our TEA-cGAN-Multi-scale model.}\label{fig:qualmultiscale}
\end{figure}

We further use our TEA-cGAN-Multi-scale model to generate higher resolution images i.e., 256 $\times$ 256. In Figure~\ref{fig:qualmultiscale}, we show TEA-cGAN-Multi-scale generated  128 $\times$ 128 and 256 $\times$ 256 resolution images side by side to show the difference. In both resolutions our model effectively disentangles text irrelevant content such as background from visual attributes that need to be changed.

\subsection{Text Interpolation}
\label{ssec:sent_interpolation}
\begin{figure}
    \centering
        \includegraphics[width=0.75\textwidth]{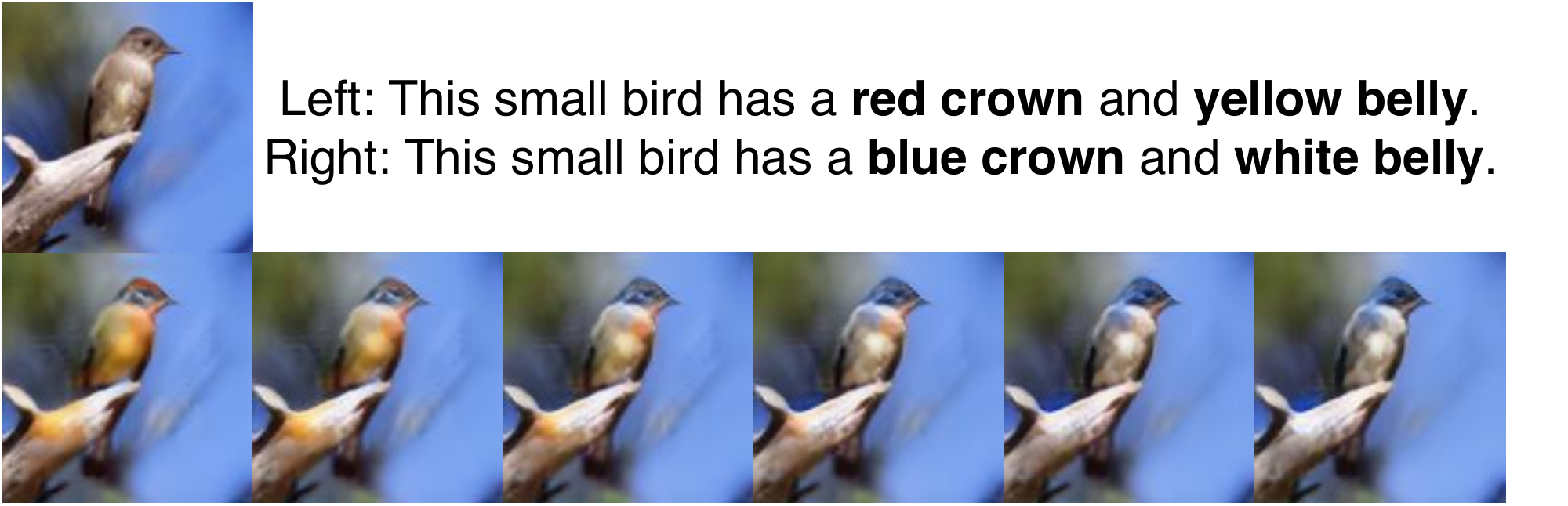}
    \caption{Text interpolation results with TEA-cGAN-Multi-scale. Upper left: the original image. Bottom: images generated according to the linear interpolation of the two sentences. All generated images have 128 $\times$ 128 resolution.}\label{fig:sentint}
\end{figure}

To understand TEA-cGAN-Multi-scale model ability to generate images without memorizing the text and is generalizable, we conducted a text interpolation experiment for the generator. The idea here is to fix the input image and select two sentences from the test set. Further, two sentences are encoded into embeddings to perform linear interpolation between them. However, in our case since we use individual word representations as opposed to a complete sentence representation, we restrict the two sentences to the same length. Figure~\ref{fig:sentint} shows that the TEA-cGAN-Multi-scale model generates images of interpolated text while preserving the contents of the original image. This validates our hypothesis of non-memorization of varied texts by inherently learning latent information useful for generalization.

\subsection{Attention Visualization}
\label{ssec:attvis}
\begin{figure}
    \centering
        \includegraphics[width=0.65\textwidth]{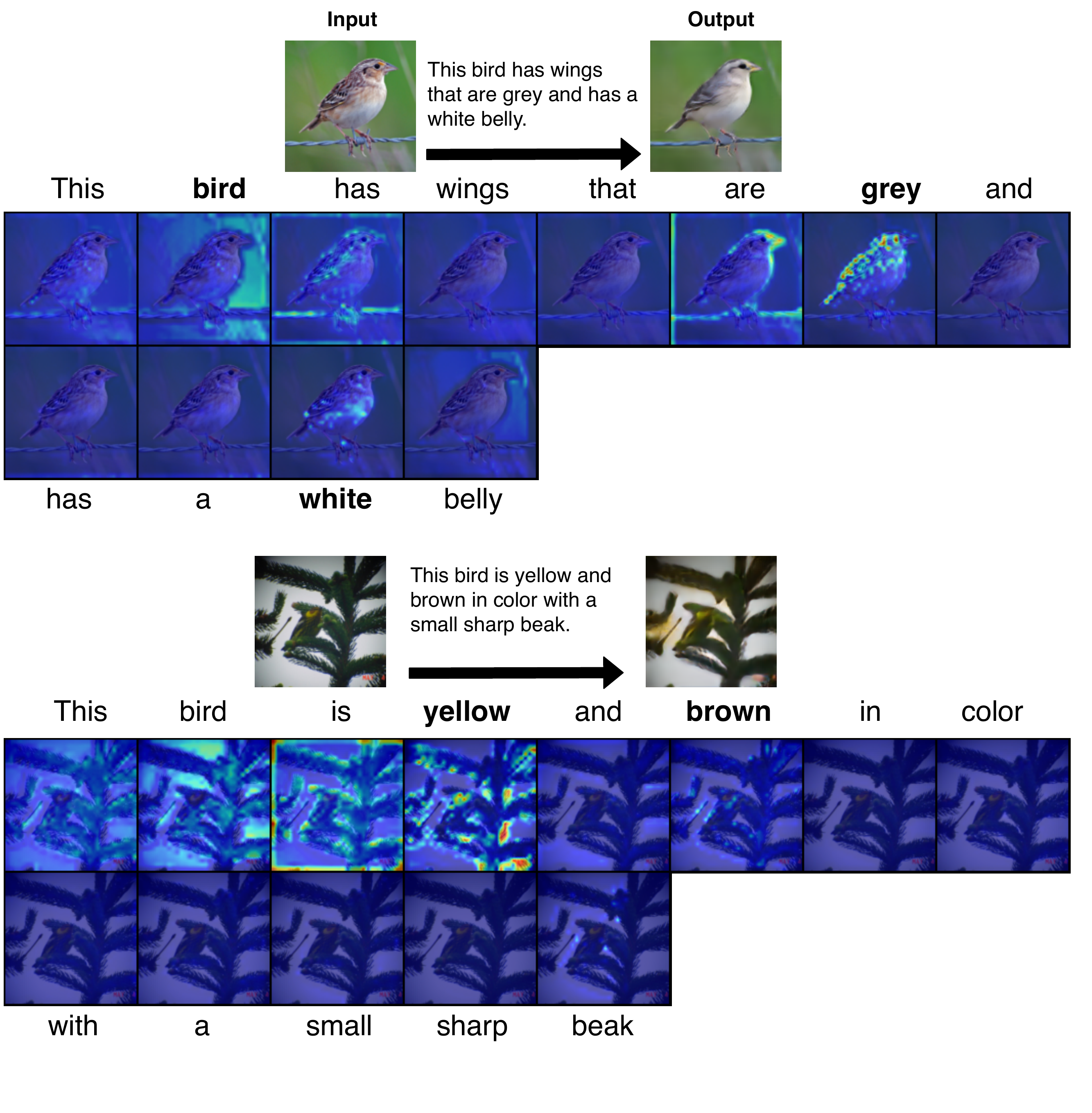}
    \caption{Attention visualization (Generator) of each word w.r.t the image with TEA-cGAN-Single-scale. Bright areas indicates locations where the model is focusing on while generating the image. \textit{Top}: Success case and \textit{Bottom}: Failure case.}\label{fig:attnvis}
\end{figure}

Images generated with TEA-cGAN-Single-scale  are visualized with heat maps in the Figure~\ref{fig:attnvis}. We observe that for some words, our model correctly attends to the corresponding locations in the images. Note that it is normal that not all words are attended to a location in the image. First, stop words do not has their corresponding locations. Second, for phrases like ``white belly'' refers to one visual attribute and the model does not need to attend to the corresponding location twice. In such cases, our model tends to distribute the attention the word describing the color (white). We see positive correlation between the attention quality and the image quality. For example, in simple scenes (Figure~\ref{fig:attnvis} top picture), our model can attend to the object itself rather than the background and generated promising images. While if the background is cluttered and the object that needs to be manipulated is invisible, then our model fails to alter visual attributes (Figure~\ref{fig:attnvis} bottom picture).

\subsection{Inception Score}
\label{subsec: inception}
Our task is similar to text-to-image generation~\cite{reed:2016, zhang2017stackgan, zhang:2018}, however we cannot use automatic evaluation measure such as inception score \cite{salimans:2016} for estimating generated image quality. This is due its inappropriateness for our task. In the following, we provide more details.

To calculate the inception score on a set of images, we basically need to apply an image classifier such as inception network \cite{salimans:2016}) on the images. Inception score is designed on an assumption that if a generated image looks realistic, the classifier should be able to classify it easily and accurately, i.e., the label distribution should have a low entropy. However, this is only true if the generator tries to generate images belonging to classes seen by the classifier. A text-to-image generation fulfills this requirement, for example, generating birds that look similar to birds in the CUB dataset. However, our generator modifies the bird and the resulting bird does not belong to any class in the CUB dataset. For example, if our generator changes the color of a crow from black to red, then the resulting bird cannot be classified. If we apply a classifier trained on the CUB dataset, it should fail to classify the modified images, not matter how realistic they look. The resulting label distribution should have higher entropy which causes a low inception score. However, it does not imply that the image quality is low.

\begin{figure*}
    \centering
        \includegraphics[width=0.65\textwidth]{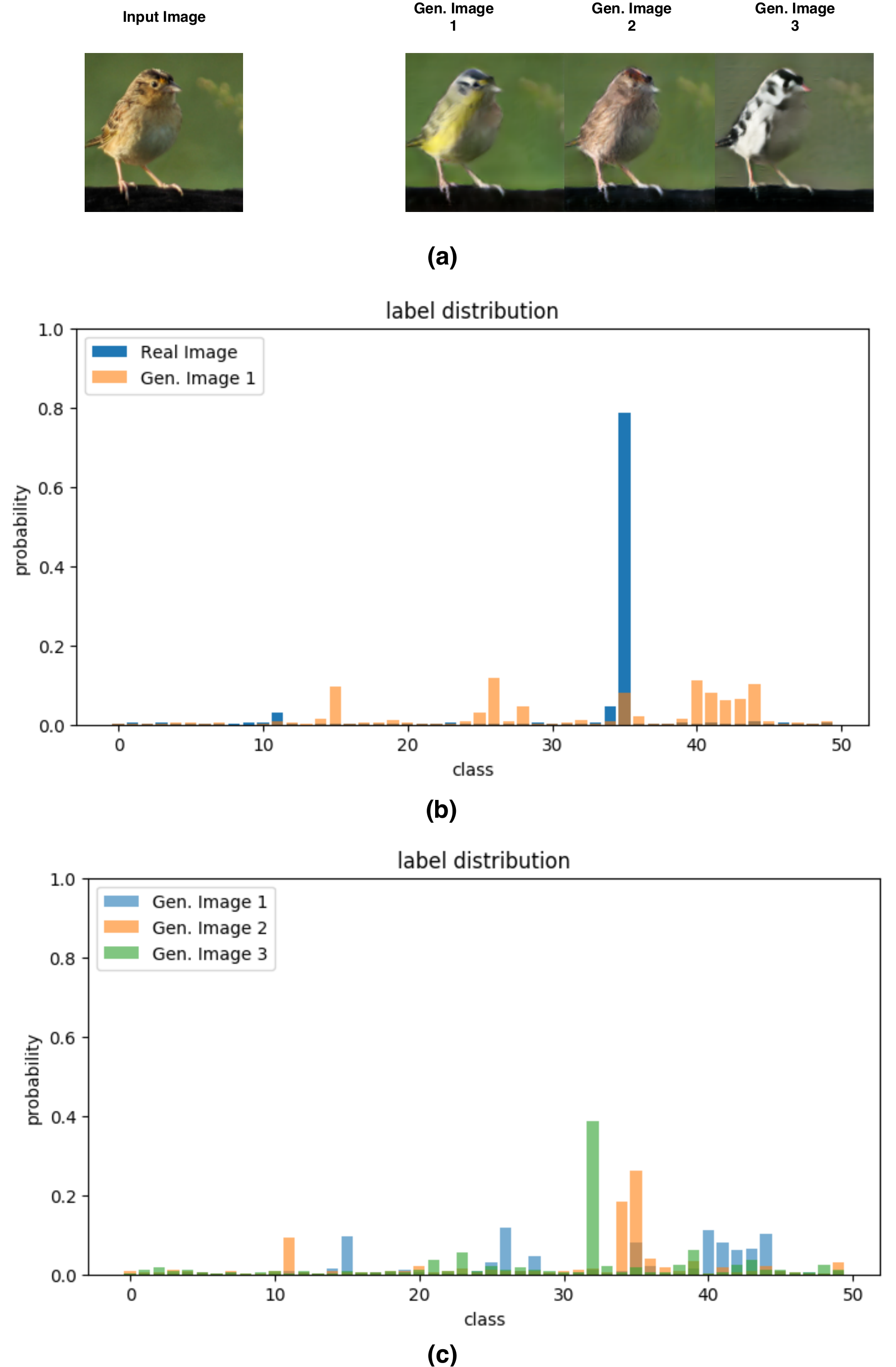}
    \caption{(a): Input image along with the three generated images according to different descriptions. (b) Label distributions of the input image and the first generated image. (c) Label distributions of the three generated images.}\label{fig:inception_res}
\end{figure*}

To verify this claim, we compare the label distributions between real and generated images. As shown in the Figure~\ref{fig:inception_res}, we use our TEA-cGAN-Multi-scale model to generate three images based on an input image according to different descriptions. We use the fine-tuned inception model \footnote{\url{https://github.com/hanzhanggit/StackGAN-inception-model}} to classify the images. We observe that the model classifies the input image correctly with high confidence ($\approx 80\%$), which results in an unimodal distribution (refer Figure~\ref{fig:inception_res} (b)). However, the model outputs a distribution that is similar to a uniform distribution when classifying the generated image. This is reasonable as this non-existent bird does not belong to any class. In the Figure~\ref{fig:inception_res} (c),  we further show that it is the same case for different descriptions. Although for a human the generated images look realistic, in our scenario the label distributions are more or less random. Therefore, the inception score is not consistent with the human perception.

\section{Related Work}
\label{sec:related}
We present related work from some of the closely aligned areas.
\subsection{Text-to-Image Generation}
\label{ssec:rwttoimgen}
Initially, alignDRAW~\cite{mansimov:2015} was introduced to iteratively draw patches on a canvas, while attending to the relevant words in the description. Further, visual concepts are translated from characters to pixels~\cite{reedlearning:2016} with a conditional GAN. It was further improved~\cite{reed:2016} by taking instructions about what content to be drawn in which location achieving high-quality image generation. 
To generate images with high resolution, several GANs are stacked together as stackGAN~\cite{zhang:2018} using the global sentence representation. This helped to generate images of different sizes. To over come the bottleneck of global-level sentence representation, attention based GAN as AttGAN~\cite{xu:2018} is introduced to capture the fine-grained details at different sub-regions of the image. It pays attention to the relevant words in the natural language description. Recently, ControlGAN~\cite{licontrollable:2019} is proposed to effectively synthesise high-quality images by controlling the parts of image generation according to natural language descriptions. Our work leverage ideas from AttGAN, however, we use it in both the generator and discriminator for manipulating image semantically. 
\subsection{Image-to-Image Translation}
\label{ssec:rwimtoimgen}
Several ideas were explored to perform image-to-image translation. There are  paired~\cite{isola:2017}, unpaired~\cite{zhuunpair:2017} and style transfer~\cite{gatys:2016} approaches proposed in the recent times based on GANs. Paired approaches that use image pairs as training examples were applied to various tasks such as generating images from sketches~\cite{sangkloy:2017}. Unpaired approaches that do not use image pairs are learned using Coupled GANs~\cite{liu:2016} and cross-modal scene networks~\cite{aytar:2017} with a weight-sharing strategy for learning a common representation. Few approaches~\cite{yi:2017} also used unsupervised techniques for image-to-image translation. Our work differs from direct image-to-image translation as we condition both image and natural language description in the generator for image generation.

\subsection{Interactive Image Manipulation}
\label{ssec:rwimmanp}
Instead of using a single natural language sentence to manipulate images, another interesting approach is to have an interactive system that generates an image iteratively. A variation of it is the image manipulation via natural language  dialogue~\cite{cheng:2018}.
\section{Conclusion and Future Work}
In this paper, we proposed TEA-cGAN which can manipulate images with a natural language description. We created two different scales of feature aggregation in the generator by leveraging attention. We found that it was helpful to find the relevant contents in the original image according to the descriptions in a more fine-grained manner. We showed in the experiments that our approach outperforms existing methods both quantitatively and qualitatively on 128 $\times$ 128 resolution and even generates higher resolution images i.e., 256 $\times$ 256 for richer experience. In future, we would like to alter images which contain multiple objects per image.

\section{Acknowledgements}
Aditya Mogadala is supported by the German Research Foundation (DFG) as part of SFB1102.

\bibliographystyle{cas-model2-names}
\bibliography{cas-sc-template}

\appendix
\newpage
\section{Additional Results}
\begin{figure}
    \centering
        \includegraphics[width=0.85\textwidth]{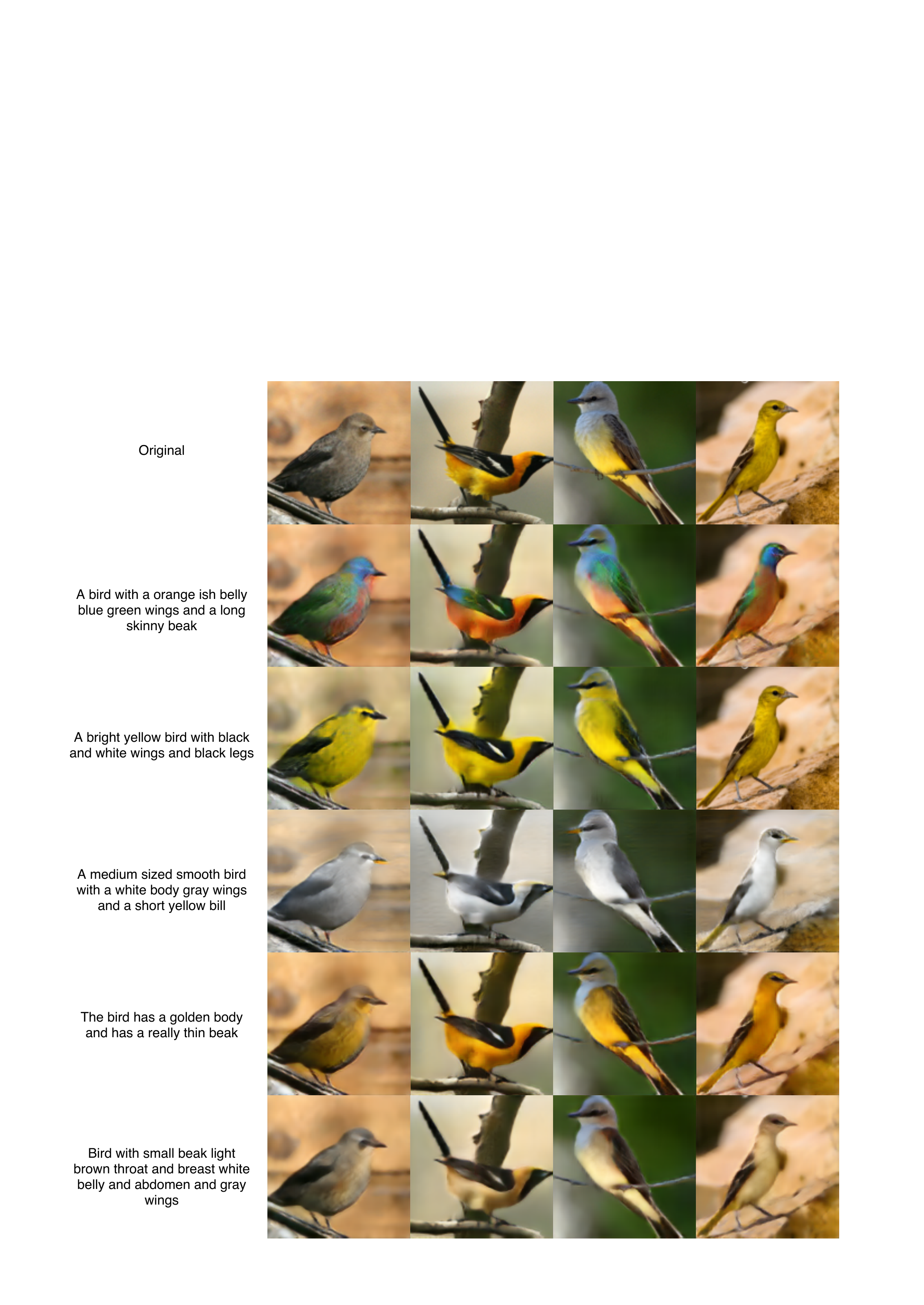}
    \caption{More generated samples with 128 $\times$ 128 resolution using TEA-cGAN-Multi-scale}\label{fig:Apx_Ex1}
\end{figure}

\begin{figure}
    \centering
        \includegraphics[width=0.85\textwidth]{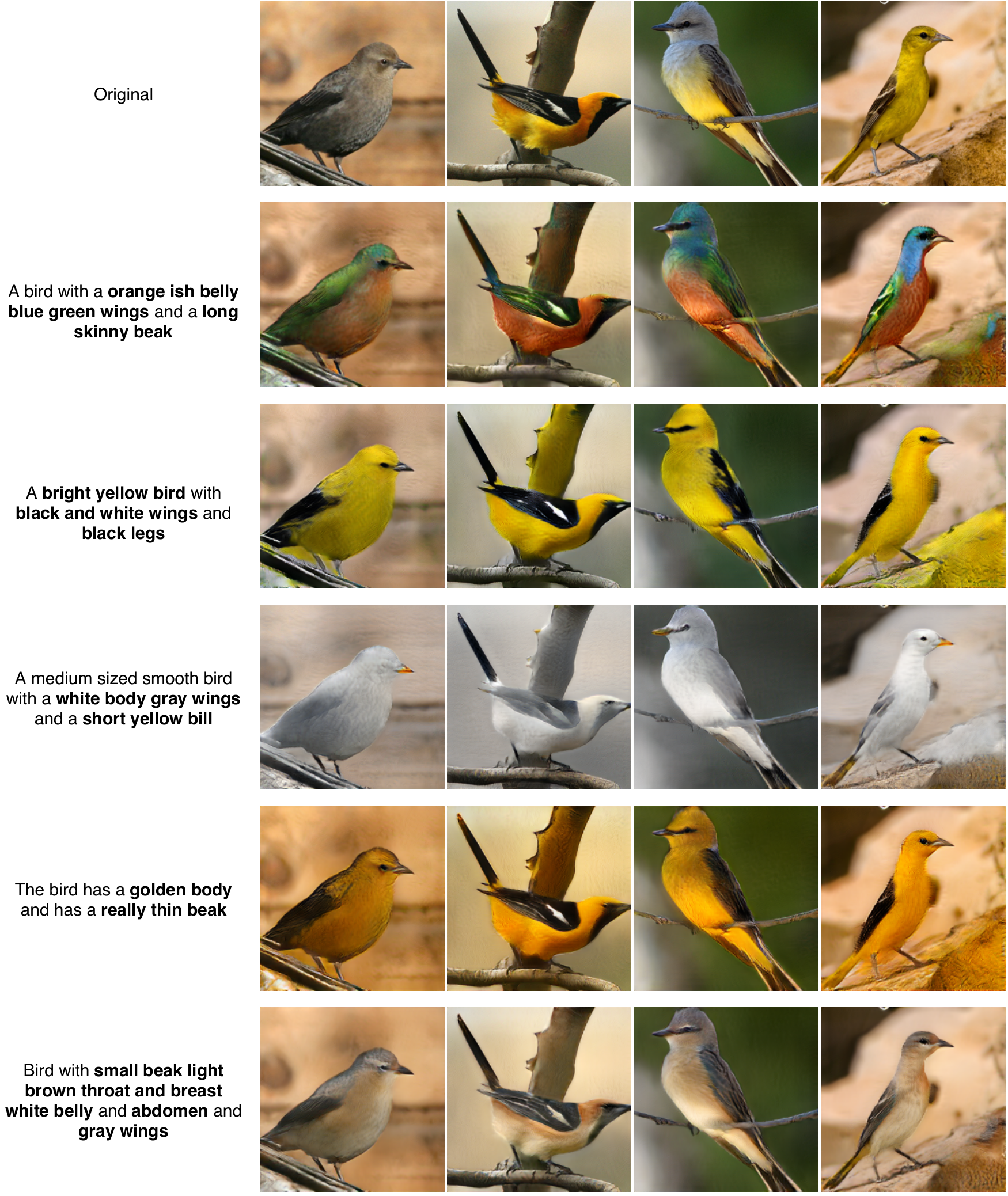}
    \caption{More generated samples with 256 $\times$ 256 resolution using TEA-cGAN-Multi-scale}\label{fig:Apx_Ex2}
\end{figure}

We show additional qualitative results of the generated samples with different resolutions by our TEA-cGAN-Multi-scale in Figure~\ref{fig:Apx_Ex1} and Figure~\ref{fig:Apx_Ex2}.

\end{document}